# Modeling failure priors and persistence in model-based diagnosis

**Sampath Srinivas**
Computer Science Department
Stanford University
Stanford, CA 94305
srinivas@cs.stanford.edu

## Abstract

Probabilistic model-based diagnosis computes the posterior probabilities of failure of components from the prior probabilities of component failure and observations of system behavior. One problem with this method is that such priors are almost never directly available. One of the reasons is that the prior probability estimates include an implicit notion of a time interval over which they are specified – for example, if the probability of failure of a component is 0.05, is this over the period of a day or is this over a week? A second problem facing probabilistic model-based diagnosis is the modeling of persistence. Say we have an observation about a system at time $t_1$ and then another observation at a later time $t_2$. To compute posterior probabilities that take into account both the observations, we need some model of how the state of the system changes from time $t_1$ to $t_2$. In this paper, we address these problems using techniques from Reliability theory. We show how to compute the failure prior of a component from an empirical measure of its reliability – the Mean Time Between Failure (MTBF). We also develop a scheme to model persistence when handling multiple time tagged observations.

## 1 INTRODUCTION

Model-based diagnosis computes the set of possible explanations consistent with a physical system's observed behavior [Hamscher *et al*, 1992]. A system consists of a set of interconnected components. Each component has a *mode* associated with it. The mode is a state variable which takes one of a fixed number of values. In the simplest case, the mode variable may take two states, **ok** and **broken**. The **ok** state denotes a mode where the component is working as it should while the **broken** state denotes a mode where the component is behaving abnormally. We have available some (partial) characterization of the the components input-output behavior. That is, given the state of the mode variable of the component and an input value we have some information about what the output might be. In the general case, this information amounts to specifying a set of possible values of the output for each mode-input combination. These component characterizations and the interconnections of the components constitute the behavior *model* of the system.

An *observation* consists of a set of variable-value pairs. Usually, the observation consists of the values of the input variables of the system and the values of some of the intermediate and output variables of the system. A *candidate* is an assignment of a state to each of the system's mode variables. So for example, say a simple system has two components $A$ and $B$, with corresponding mode variables $M_A$ and $M_B$. Each of the mode variables can take two states, **ok** and **broken**. For this system, one possible candidate is $\langle M_A = \mathbf{ok}, M_B = \mathbf{broken}\rangle$.

The process of *diagnosis* consists of computing the set of *candidates* which are consistent with the observation and model. One drawback of strictly logical approaches to diagnosis is that there is no notion of plausibility. For non-trivial systems, the set of consistent candidates can be very large. A mechanism to order the candidates by likelihood is required.

Probabilistic model-based diagnosis introduces a prior distribution over each component's mode variable. Each component's failure is assumed independent of the failure of other components. For example, the component $A$ might have the prior $P(M_A = \mathbf{ok}) = 0.995$ and $P(M_A = \mathbf{broken}) = 0.005$. The diagnosis problem is now defined as computing a posterior probability distribution over the set of possible candidates. Let $C$ be a state variable which ranges over all possible candidates and $\Omega$ be an observation. The diagnosis problem now amounts to computing $P(C|\Omega)$ [Geffner & Pearl, 1987; DeKleer & Williams, 1989; Poole, 1991; Srinivas, 1994].

One of the drawbacks of probabilistic diagnosis is that the specification of the prior probabilities of failures of the components is difficult. The probabilities are not usually directly available from an expert. One of the reasons for this is that there is an implicit notion of time in the prior. For example, say a system has been running for a week at the time of diagnosis. Say we assess the *prior* probability of failure of a component $A$ at the time we carry out our diagnosis and find that $P(M_A = \textbf{broken}) = p_1$. Now say the system has been running for a month and we assess $P(M_A = \textbf{broken})$ again. Intuitively, the probability in the latter case should be different than $p_1$. Furthermore, it should be larger – the longer the component has been on-line the higher the chances that it has failed.

A related problem occurs when doing diagnosis with multiple observations, where each observation occurs at a different time. When performing diagnosis with multiple observations we need some model of the persistence of the state of the system between the observations. For example, say we have an observation $\Omega[t_1]$ at time $t_1$ and an observation $\Omega[t_2]$ at time $t_2$. Let the candidate state variable at time $t_1$ be $C[t_1]$ and that at time $t_2$ be $C[t_2]$. We will use $c[t_1]$ and $c[t_2]$ to denote states of $C[t_1]$ and $C[t_2]$. The goal of diagnosis is to compute the distribution $P(C[t_2] \mid \Omega[t_1], \Omega[t_2])$. This distribution can be computed as:

$$\begin{aligned}
P(C[t_2] = c[t_2] &\mid \Omega[t_1], \Omega[t_2]) \propto \\
&P(C[t_2] = c[t_2], \Omega[t_1], \Omega[t_2]) \\
\propto \quad &P(\Omega[t_2] \mid C[t_2] = c[t_2], \Omega[t_1]) \\
&\times P(C[t_2] = C[t_2] \mid \Omega[t_1]) \\
= \quad &P(\Omega[t_2] \mid C[t_2] = c[t_2]) \\
&\times P(C[t_2] = c[t_2] \mid \Omega[t_1]) \quad (1)
\end{aligned}$$

The last step follows because knowing the state $C[t_2]$ of the system renders the observation $\Omega[t_2]$ independent of the history. The probability $P(\Omega[t_2] \mid C[t_2] = c[t_2])$ can be calculated from the system model. In the above equation we also need the distribution $P(C[t_2] \mid \Omega[t_1])$, i.e., an estimate of the state at time $t_2$ given the observation at time $t_1$. This can be computed as follows:

$$\begin{aligned}
P(C[t_2] = c[t_2] &\mid \Omega[t_1]) = \\
\Sigma_{c[t_1]} &P(C[t_2] = c[t_2] \mid C[t_1] = c[t_1], \Omega[t_1]) \\
&\times P(C[t_1] = c[t_1] \mid \Omega[t_1]) \\
= \quad \Sigma_{c[t_1]} &P(C[t_2] = c[t_2] \mid C[t_1] = c[t_1]) \\
&\times P(C[t_1] = c[t_1] \mid \Omega[t_1]) \quad (2)
\end{aligned}$$

The last step follows from a Markov assumption that the system state $C[t_2]$ is rendered independent of all history upto time $t_1$ when the state $C[t_1]$ at time $t_1$ is known. In the above equation, $P(C[t_1] \mid \Omega[t_1])$ is the posterior over the state of the system at time $t_1$. This can be computed easily as:

$$\begin{aligned}
P(C[t_1] = c[t_1] &\mid \Omega[t_1]) \propto \\
&P(\Omega[t_1] \mid C[t_1] = c[t_1]) \times P(C[t_1] = c[t_1])
\end{aligned}$$

The first term in the RHS of the above equation can be computed from the system model. The second term is the prior probability of the candidate. This can be computed as a product from the prior distribution over the mode of the individual components.

We see that the distribution $P(C[t_2] \mid C[t_1])$ in Equation 2 is still unspecified. This distribution quantifies the persistence of the state of the system. For example, if the system never changed state spontaneously then we would have:

$$P(C[t_2] = c[t_2] \mid C[t_1] = c[t_1]) = \begin{cases} 1 & \text{if } c[t_1] = c[t_2] \\ 0 & \text{if } c[t_1] \neq c[t_2] \end{cases}$$

However, a more practical situation would be one where components can fail on-line. If we had a model of how the components failed when they were on-line, this could be used to compute the distribution $P(C[t_2] \mid C[t_1])$.

In this paper we address the problems of specifying failure priors and modeling persistence using techniques from reliability theory. Reliability theory gives us a model that relates the probability of component failure and the age of the component. We use this model to compute the prior probability of failure from the uptime of component and the Mean Time Between Failure (MTBF). The MTBF is a summary measure of the reliability of a device and is often available with manufacturer specifications. We also use the component failure model to compute the conditional distribution over the state of a component at time $t_2$ given the state at time $t_1$. This conditional distribution is used to model persistence.

## 2 RELIABILITY MODELS

Reliability theory offers empirically validated models for the failure process of devices. The failure process of a device relates the probability of failure of the device to time (for example, see [Tsokos & Shimi, 1977]). We now describe one standard model of the failure process.

Consider a device $A$. It has two states – **ok** and **broken**. It is initially in the **ok** state when it is brought on-line. At some point in time it transitions into the **broken** state. Once it is broken, it stays broken. It is assumed that the failure of $A$, i.e., the transition of $A$ from the **ok** to **broken** state, occurs due to random unmodeled events which occur with a probability that is constant over time. For example, an electronic component may fail due to surges in the power supply – these surges may occur randomly with a uniform probability.

The modeling assumption is interpreted as follows: Given that the device has not failed at time $t$, the conditional probability that it will fail in the interval $[t, t+dt]$ is proportional to the length of the interval. The probability is thus given by $\lambda dt$ where $\lambda$ is a proportionality constant. Say we model the failure of $A$ with a continuous real variable $X$. That is, "$X \leq t$" denotes the event that $A$ fails in the interval $[0, t]$. Say $F(t)$ is the cumulative distribution of $X$, i.e., $F(t) = P(0 \leq X \leq t)$. $F(t)$ is the probability that $A$ fails in the time interval $[0, t]$.

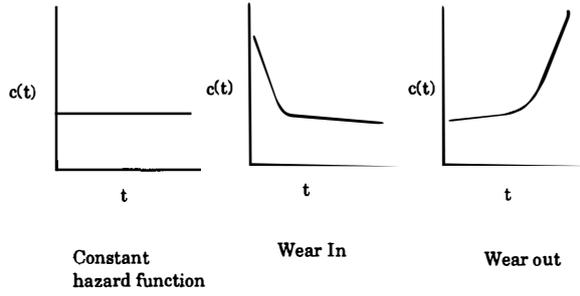

Figure 1: Different choices for the hazard function.

From the modeling assumption, we have:

$$P(t < X \leq t + dt \mid X > t) = \lambda dt \quad (3)$$
$$\frac{F(t + dt) - F(t)}{1 - F(t)} = \lambda dt$$
$$\frac{dF(t)}{dt} = \lambda(1 - F(t))$$
$$F(t) = 1 - e^{-\lambda t}$$
$$\text{(Since F(0) = 0)}$$

We see that the probability of $A$ failing before time $t$ is small when $t$ is small and converges to 1 as $t$ tends to infinity.

In general, instead of assuming a constant $\lambda$ in the equation above, we might have a time varying conditional density $c(t)$, called the *hazard function*. This allows us to model various different degradation phenomena in the device. For example, if we want to model a situation where a component has high probability of failure early in its life (called "wear in") we might choose $c(t)$ to be high for small values of $t$. If we want to model wear out (and hence, increased chance of failure) of the device as it grows older we might choose $c(t)$ to be high as $t$ becomes large (See Fig 1). The differential equation Equation 3 can be solved for the appropriate choice of hazard function to give the corresponding density function.

The expectation of the variable $X$ is called the Mean Time between Failures. In the case of $c(t) = \lambda$, the MTBF can be shown to be $\frac{1}{\lambda}$.

### 2.1 COMPUTING PRIOR PROBABILITIES OF FAILURE

Using the above model, we can specify more precisely what we mean by the prior distribution over the mode of a component $A$. The probability that $A$ is broken is a function of time. We will denote this probability at time $t$ by $P(M_A[t] = \textbf{broken})$. Say that the last time we knew that $A$ was certainly not broken was $t_{ok}$. Usually, $t_{ok}$ is the time that $A$ first went on-line.

We see that $P(M_A[t] = \textbf{broken})$ is the probability that $A$ failed in the interval $[t_{ok}, t]$. Say we assume a constant hazard function. We see that:

$$\begin{aligned} P(M_A = \textbf{broken}) &= P(t_{ok} < X \leq t \mid X > t_{ok}) \\ &= \frac{F(t) - F(t_{ok})}{1 - F(t_{ok})} \\ &= 1 - e^{-\lambda(t - t_{ok})} \quad (4) \end{aligned}$$

The above equation allows us to directly relate MTBFs to the prior probabilities that we need.

Note that when we assume $c(t) = \lambda$, $P(M_A = \textbf{broken})$ is only a function of $t - t_{ok}$. That is, all we need to know is the length of the interval from the last time at which we were sure that $A$ was not broken. However, if we assume more complicated hazard functions $c(t)$, we would also need to know when $A$ first went on-line. This is because we will measure $t$ and $t_{ok}$ using that as the start time. Hence in general, we will find that $P(M_A = \textbf{broken}) = f(t, t_{ok})$ where $f$ is some function whose form depends on the choice we make for the hazard function $c(t)$.

### 2.2 MODELING PERSISTENCE

Consider the example outlined in Section 1. We saw there that some persistence model was necessary to compute $P(C[t_2] \mid C[t_1])$. The reliability model described above gives us a method to compute this distribution.

Say the system has $n$ components, $C_1, C_2, \ldots, C_n$. The mode variable corresponding to component $C_i$ is $M_i$. We assume that the failure processes of the individual components are independent. Thus, given the state of $C_i$ at time $t_1$, we can compute a conditional probability distribution over the state at time $t_2$ without regard to any of the other components. If we assume a constant hazard function $\lambda_i$ for $C_i$, this distribution is given by Table 1.

The state $c[t]$ of the variable $C[t]$ consists of a state assignment to each of the mode variables $M_i$. Denoting the state of $M_i[t]$ by $m_i[t]$ we have $c[t] = \langle M_1[t] = m_1[t], M_2[t] = m_2[t], \ldots, M_n[t] = m_n[t] \rangle$. From the independence assumption of the failure processes we have:

$$\begin{aligned} P(C[t_2] = c[t_2] \mid C[t_1] = c[t_1]) = \\ P(M_1[t_2] = m_1[t_2], \ldots, M_n[t_2] = m_n[t_2] \mid \\ M_1[t_1] = m_1[t_1], \ldots, M_n[t_1] = m_n[t_1]) \\ = \Pi_{1 \leq i \leq n} P(M_i[t_2] = m_i[t_2] \mid M_i[t_1] = m_i[t_1]) \end{aligned}$$

Each of $P(M_i[t_2] = m_i[t_2] \mid M_i[t_1] = m_i[t_1])$ can be computed as shown in Table 1. Thus, the failure processes of the individual components and an assumption about their independence gives us a model for the persistence of the system state. The distribution $P(C[t_2] \mid C[t_1])$ can be computed using $t_2$, $t_1$ and the MTBFs of the individual components $M_i$.

Table 1: Persistence of $M_i$ from $t_1$ to $t_2$.

| $P(M_i[t_2] \mid M_i[t_1])$ | $M_i[t_1] = $ ok | $M_i[t_1] = $ broken |
|---|---|---|
| $M_i[t_2] = $ ok | $e^{-\lambda_i(t_2-t_1)}$ | 0 |
| $M_i[t_2] = $ broken | $1 - e^{-\lambda_i(t_2-t_1)}$ | 1 |

## 3 USING RELIABILITY MODELS IN DIAGNOSIS

We illustrate the use of reliability models in model-based diagnosis with some examples. We will first describe a system and a decision theoretic repair model for the system. We will then describe two different scenarios with this system.

The first scenario demonstrates how diagnosis/repair depends crucially on *when* an observation is made. The observation in this case is a "nothing wrong" observation. That is, the observation does not indicate any discrepancy. We show that if this observation was made when the system was new then the probability of everything being ok is very large (as we would expect) and the optimal decision is to do nothing (again, as we would expect). However, if the same observation is made instead when the system is older, the posterior probability that everything is ok decreases substantially. In this case, the optimal decision is to actually replace two of the components. This decision can be seen as an example of preventive maintenance.

The second scenario demonstrates the need for persistence models in the presence of multiple observations and actions. We first make an observation at time $t_1$. The corresponding best action is computed and executed and is found to fix the anomaly exhibited by the system. Now, at a later time $t_2$, the same anomalous observation occurs. A persistence model for the component states is required to compute the posterior probabilities and the optimal decision after the second observation. Though the observation at time $t_2$ and time $t_1$ are the same, the posterior distribution and the optimal decision are very different.

### 3.1 THE EXAMPLE SYSTEM

Consider the digital circuit shown in Fig 2. $A$ is an And gate, $O$ is an Or gate and $X$ is an Xor gate. Each of the gates can be **ok** or **broken**. When a gate is **ok** it works as it is supposed to. When it is **broken** we need a fault model. For the purposes of this example we will assume that we have a fully specified fault model for each gate. Specifically, we will assume *stuck-at-0* fault models for $A$ and $O$ and a *stuck-at-1* fault model for $X$. That is, when $A$ (or $O$) is in the **broken** state the output of the gate is 0 irrespective of the input. Similarly, when $X$ is **broken**, its output is 1 irrespective of the input. Say the MTBF of the And gate $A$ is 100 hours, that of the Or gate $O$ is 250

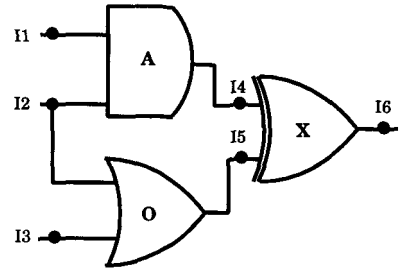

Figure 2: The example circuit.

Table 2: Repair cost functions.

| $L_A(m,d)$ | | | $L_O(m,d)$ | | |
|---|---|---|---|---|---|
| b | fix | $2 | b | fix | $3 |
| b | dont | $8 | b | dont | $12 |
| ok | fix | $2 | ok | fix | $3 |
| ok | dont | $0 | ok | dont | $0 |

| $L_X(m,d)$ | | |
|---|---|---|
| b | fix | $4 |
| b | dont | $14 |
| ok | fix | $4 |
| ok | dont | $0 |

hours and that of the Xor gate $X$ is 350 hours.

We also have a cost model that describes the repair costs of the system. We associate a repair decision with each component. The decision alternatives are **fix** and **dont-fix**. We describe the repair cost for each component with a cost function whose arguments are its mode state and the repair alternative. The cost functions for each of the gates is shown in Table 2. The system repair cost is assumed to be the sum of the repair costs of the components. That is, given a state for each mode variable of the system and a repair decision alternative for each component, we can sum the cost of the repair decision alternative for each component to get the repair cost function $L_S$ for the entire system. The arguments to this cost function are a candidate (i.e., state assignment to each component of the system) and a *composite* decision. A composite decision consists of a choice of a repair alternative for each component in the system.

The cost function $L$ for each component is quite intuitively assessed. Consider the case where the component $A$ is **ok** and the decision is **dont-fix**. We use this as the reference situation and assign it a cost of $0. Hence $L_A(\text{ok},\text{dont-fix}) = \$0$. When we do decide to **fix**, say that we throw the old component away and replace it with a new one irrespective of what its actual state is. The cost of implementing the **fix** decision summarizes the temporary downtime cost required to bring the system down to change the component *and* the cost of the new component. Say we determine this cost to be $2. Hence $L_A(\text{broken},\text{fix}) = L_A(\text{ok},\text{fix})$

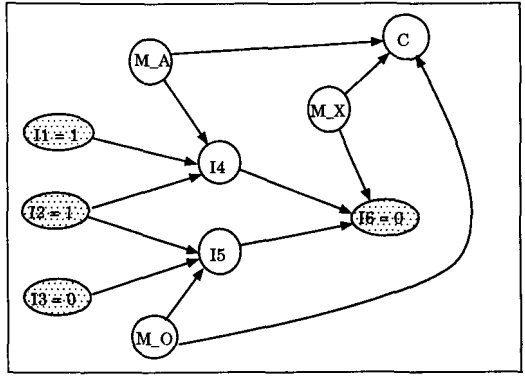

Figure 3: Bayesian network for Scenario 1.

= $2. Finally, we assign a cost to the case where the gate is **broken** but we choose **dont-fix**. This cost basically summarizes the cost of system downtime as a result of this non-operational component. Say we determine the cost to be $10. Hence $L_A(\textbf{broken},\textbf{dont-fix}) = \$10$. There is an implicit time horizon over which these costs are being measured – say it is 1 hour in this example.

In our framework, diagnosis involves computing the posterior distribution over the candidate state variable given observations. We will later see how these calculations can be made by translating the system model into a Bayesian network.

Repair involves choosing the optimal composite decision for the entire system given the posterior distribution over the candidates. The optimal decision $d_{opt}$ is the one which results in the least expected cost. That is:

$$d_{opt} = min_d(\Sigma_c P(C = c|\Omega) L_S(D = d, C = c))$$

In this equation, $d$ ranges over all possible composite decisions and $c$ ranges over all possible candidates.

### 3.2 SCENARIO 1: EFFECT OF TIME ON DIAGNOSIS/REPAIR

Consider the system of Fig 2. Say that it has been been up for 10 hours and the observation $\Omega = \langle I_1 = 1, I_2 = 1, I_3 = 0, I_6 = 0\rangle$ is recorded. Note that $\Omega$ seems to suggest that there is no problem with the system since the inputs generate the expected output. We assume that the time is counted from the time the system was new. We will further assume that when the system was new every component was certainly **ok**.

The system model can be translated into a Bayesian network as shown in Fig 3 (see [Srinivas, 1994] for details). The node $C$ has 8 states, one corresponding to each combination of states of $M_A$, $M_O$ and $M_X$. Let $m_A$ be a state of $M_A$. We define $m_O$ and $m_X$ similarly. The conditional distribution of $C$ given its parents is just a deterministic distribution which maps each state combination of its parents to the corresponding state of $C$. That is, we have $P(C = \langle m_A, m_O, m_X \rangle \mid M_A = m_A, M_O = m_O, M_X = m_X) = 1$.

Table 3: System up 10 hours, $\Omega$ observed.

| Failure priors | | | |
|---|---|---|---|
| Comp. | Uptime | MTBF | $P(M = \textbf{b})$ |
| A | 10 | 100 | 0.0952 |
| O | 10 | 250 | 0.0392 |
| X | 10 | 350 | 0.0282 |

| Posterior $P(C|\Omega)$ | | | |
|---|---|---|---|
| A | O | X | |
| ok | ok | ok | 0.9957 |
| b | b | ok | 0.0043 |
| ok | ok | b | 0.0000 |
| ok | b | ok | 0.0000 |
| ok | b | b | 0.0000 |
| b | ok | ok | 0.0000 |
| b | ok | b | 0.0000 |
| b | b | b | 0.0000 |

| Expected cost | | | |
|---|---|---|---|
| $D_A$ | $D_O$ | $D_X$ | $ |
| dont | dont | dont | 0.0855 |
| fix | dont | dont | 2.0513 |
| dont | fix | dont | 3.0342 |
| dont | dont | fix | 4.0855 |
| fix | fix | dont | 5.0000 |
| fix | dont | fix | 6.0513 |
| dont | fix | fix | 7.0342 |
| fix | fix | fix | 9.0000 |

Table 4: System up 90 hours, $\Omega$ observed.

| Failure priors | | | |
|---|---|---|---|
| Comp. | Uptime | MTBF | $P(M = \textbf{b})$ |
| A | 90 | 100 | 0.5934 |
| O | 90 | 250 | 0.3023 |
| X | 90 | 350 | 0.2267 |

| Posterior $P(C|\Omega)$ | | | |
|---|---|---|---|
| A | O | X | |
| ok | ok | ok | 0.6126 |
| b | b | ok | 0.3874 |
| ok | ok | b | 0.0000 |
| ok | b | ok | 0.0000 |
| ok | b | b | 0.0000 |
| b | ok | ok | 0.0000 |
| b | ok | b | 0.0000 |
| b | b | b | 0.0000 |

| Expected cost | | | |
|---|---|---|---|
| $D_A$ | $D_O$ | $D_X$ | $ |
| fix | fix | dont | 5.0000 |
| dont | fix | dont | 6.0995 |
| fix | dont | dont | 6.6493 |
| dont | dont | dont | 7.7488 |
| fix | fix | fix | 9.0000 |
| dont | fix | fix | 10.0995 |
| fix | dont | fix | 10.6493 |
| dont | dont | fix | 11.7488 |

We now perform diagnosis using the observation $\Omega$ and the time of observation (viz, 10 hours). The prior probability of failure for each component is calculated on the basis of it being up for 10 hours. The evidence $\Omega$ is declared in the network (shown as grayed nodes) and then network inference is performed. We then look up the posterior distribution of $C$ and compute the expected costs of the decisions. These are shown in Table 3. The results confirm our intuitions, viz, the most probable diagnosis is that nothing is wrong. The optimal decision is to not fix anything.

Consider now a situation where, instead of 10 hours, the system was running for 90 hours when observation $\Omega$ was made. In this case, the prior probability of failure of each component is computed on the basis of it being up for 90 hours. We then do the diagnosis and compute the expected value of decisions for this situation. The posterior distributions and expected decision costs are shown in Table 4. In this situation, the most probable diagnosis is again that everything is ok, the same as when $\Omega$ was observed after 10 hours. However, the most probable diagnosis is much less probable. The optimal decision in this situation however, is to replace the And gate and Or gate. This is an example of preventive maintenance. The expected costs of replacing now is lower than deferring the replacement. This is because the probability of failure is rising to a critical value.

### 3.3 SCENARIO 2: MODELING PERSISTENCE

Say the system of Fig 2 has been been up for time $t_1 = 20$ hours and the observation $\Omega[t_1] = \langle I_1 = 0, I_2 = 0, I_3 = 0, I_6 = 1 \rangle$ is recorded. Note that $\Omega[t_1]$ indicates some problem with the system – if everything was working correctly the output $I_6$ would have value 0. Say we compute the posterior probabilities and the optimal decision. This computation is similar to that of the previous section. The posterior probabilities and optimal decision are shown in Table 5. The optimal decision is to fix the $XOR$ gate alone. Say we carry out this decision and then observe (immediately) that the output $I_6$ changes to 0. Thus, the observed discrepancy is fixed. Now say 20 hours more elapse and at time $t_2 = 40$ hours, we observe $\Omega[t_2] = \langle I_1 = 0, I_2 = 0, I_3 = 0, I_6 = 1 \rangle$. Note that $\Omega[t_1]$ and $\Omega[t_2]$ have the same values for the input and output variables.

The situation is represented by the dynamic Bayesian network of Fig 4. The section of the figure within the boundary marked $t_1$ represents the situation at time $t_1$. The variable $M'_X[t_1]$ represents the state of the $XOR$ gate immediately after it is replaced. Note that immediately after replacement we know that the gate is in the ok state. This is represented in the Bayesian network as evidence (grayed node). The variable $I'_6[t_1]$ represents the output immediately after the replacement action. As per our scenario, this variable is observed to have value 0.

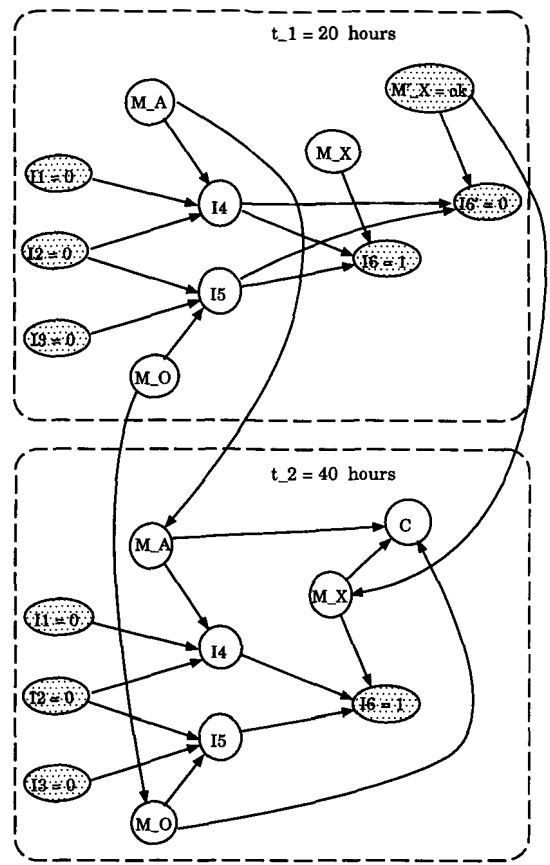

Figure 4: Bayesian network for Scenario 2.

The section of Fig 4 within the boundary marked $t_2$ represents the situation at time $t_2$. The link between $M_A[t_1]$ and $M_A[t_2]$ represents the persistence of the state of the $AND$ gate. We compute the distribution $P(M_A[t_2]|M_A[t_1])$ using Table 1. When using the table, we set $\lambda_A = \frac{1}{MTBF_A} = \frac{1}{100}$ and $t_2 - t_1 = 20$. The conditional distributions $P(M_O[t_2]|M_O[t_1])$ and $P(M_X[t_2]|M'_X[t_1])$ and are computed similarly. The anomalous observation at time $t_2$ is also entered as evidence in the Bayesian network (shown as grayed nodes).

The node $C[t_2]$ has 8 states each of which corresponds to one joint state of $M_A[t_2]$, $M_X[t_2]$ and $M_O[t_2]$. The conditional distribution of $C[t_2]$ is quantified as described in the previous section. The posterior distribution over the states of the system at time $t_2$ can be computed simply by doing inference in the Bayesian network and looking up the posterior distribution of $C[t_2]$. The posterior distribution at time $t_2$ and the corresponding decision costs are shown in Table 6. The optimal decision now is to replace both the $XOR$ and the $AND$ gate.

Table 5: Posteriors and costs at time $t_1$ (before repair).

| Posterior $P(C\|\Omega)$ | | | |
|---|---|---|---|
| A | O | X | |
| ok | ok | b | 0.7558 |
| b | ok | b | 0.1673 |
| ok | b | b | 0.0629 |
| b | b | b | 0.0139 |
| b | b | ok | 0.0000 |
| b | ok | ok | 0.0000 |
| ok | b | ok | 0.0000 |
| ok | ok | ok | 0.0000 |

| Expected cost | | | |
|---|---|---|---|
| $D_A$ | $D_O$ | $D_X$ | $ |
| dont | dont | fix | 6.3728 |
| fix | dont | fix | 6.9226 |
| dont | fix | fix | 8.4502 |
| fix | fix | fix | 9.0000 |
| dont | dont | dont | 16.3728 |
| fix | dont | dont | 16.9226 |
| dont | fix | dont | 18.4502 |
| fix | fix | dont | 19.0000 |

Table 6: Posteriors and costs at time $t_2$.

| Posterior $P(C\|\Omega)$ | | | |
|---|---|---|---|
| A | O | X | |
| ok | ok | b | 0.5712 |
| b | ok | b | 0.2809 |
| ok | b | b | 0.0991 |
| b | b | b | 0.0487 |
| b | b | ok | 0.0000 |
| b | ok | ok | 0.0000 |
| ok | b | ok | 0.0000 |
| ok | ok | ok | 0.0000 |

| Expected cost | | | |
|---|---|---|---|
| $D_A$ | $D_O$ | $D_X$ | $ |
| fix | dont | fix | 7.7743 |
| dont | dont | fix | 8.4117 |
| fix | fix | fix | 9.0000 |
| dont | fix | fix | 9.6374 |
| fix | dont | dont | 17.7743 |
| dont | dont | dont | 18.4117 |
| fix | fix | dont | 19.0000 |
| dont | fix | dont | 19.6374 |

Superficially, the situation at time $t_1$ and $t_2$ seem to be quite similar, viz, that (a) the system has been working for 20 hours since "things were ok" (b) the observation $\langle I_1 = 0, I_2 = 0, I_3 = 0, I_6 = 1\rangle$ is made. However, the situations are actually quite different. At time $t_2$ we have to account for the (a) the observations and actions at time $t_1$ and (b) possible failures of components between time $t_1$ and $t_2$.

The persistence model for the system allows us to incorporate this information when computing the posterior distribution at time $t_2$. Note that the Bayesian network inference is implicitly carrying out the computation of $P(C[t_2] \mid C[t_1])$ (discussed in Section 1).

## 4 DISCUSSION

Modeling persistence in diagnosis has been of interest both in the model-based diagnosis community and in the Bayesian network community. Early work in the model-based diagnosis community has handled persistence only to the extent of assuming that in the case of multiple observations the state of each component stays the same across all observations. This corresponds, in our framework, to the situation where multiple observations are made very close in time.

[Portinale, 1992] addresses the problem of temporal evolution of state in model-based diagnosis. The evolution is modeled as a discrete time Markov chain. The state transition matrix is assumed to come from reliability measures of components. A uniform initial prior on all possible world states is used. A definition of a temporal diagnosis is proposed that generalizes the definition of a diagnosis in a static system. A method of eliminating very unlikely diagnoses is proposed. Our approach has a similar motivation – viz, to use models of component failure processes to model change of system state. However, our approach is significantly more general. We show that the problem of specification of priors can be addressed within the same framework. The priors are computed directly from the time at which diagnosis takes place and the MTBFs of the components. Our approach also allows modeling of effects of repair actions (as in the example of Section 3.3). Time is considered to be continuous. Finally, rather than eliminating unlikely diagnoses, we compute posterior probabilities over candidates. This is necessary if a decision theoretic scheme for choosing actions is to be used.

The model of persistence developed in this paper is closely related to work in modeling persistence in Bayesian networks. [Dean & Kanazawa, 1989] propose a modeling framework for persistence and change using temporal probabilistic networks. A Markov assumption is employed. The temporal probabilistic network is quantified by specifying the conditional probability of each proposition in the network given the state of the same proposition at the immediately preceding time point and the states of the causes of the proposition at the previous time point. An exponentially decaying survivor function is proposed to model the probability of a proposition being true if it was true at the previous time and none of the causes that make it untrue are active. The survivor function accounts for unmodeled causes that might change the state of the proposition.

The model presented in this paper can be viewed as a specific instantiation of the general framework suggested by Dean and Kanazawa. We assume that the only system state that persists across time is the mode of the variables and that the failure processes of the components are independent. This allows us to model persistence of system state by simply modeling the persistence of each component individually. The persistence model for each component falls directly out of the physical process of failure postulated by the reliability model. This persistence model is an exponential decay function.

The modeling of action (and hence of persistence) in causal probabilistic networks has been a field of active interest [Balke & Pearl, 1994; Darwiche & Goldzmidt, 1994; Heckerman & Shachter, 1994; Pearl, 1994]. When reasoning about actions which occur over separated points in time, one also has to account for possible changes in system state occurring due to unmodeled events. This paper advances a specific method for doing so in the domain of diagnosis of physical systems.

Our method for modeling persistence has been developed assuming that the behavior of each component is deterministic, both when it is in the **ok** state and and when it is in the **broken** state. This allows us to consider only the mode variables of the components as the persisting state of the system. This is because the joint state of the mode variables completely determines the behavior of the system. In the situation where components are not deterministic[1], as a first approximation, one might use the same modeling scheme as the one presented in this paper. This would mean, in effect, that we assume that each component "resamples" to compute its output in each observation. Thus, the same component with the same input could possibly have two different outputs in two observations (if the component were in the broken state). If this approximation is inaccurate, modeling extensions along the lines suggested by Darwiche and Goldszmidt can be incorporated.

The simplistic decision theoretic repair model set up in this paper is for illustration purposes only. More sophisticated and realistic repair schemes (see [Heckerman et al, 1995], [Srinivas, 1995]) have been suggested in the literature. These can be used in concert with the persistence model developed in this paper.

---

[1] For example, we might assume that all outputs are equally likely if the component is **broken** in the case that no fault model is available.